\pdfoutput=1

\documentclass[11pt,a4paper]{article}
\usepackage[table]{xcolor}
\usepackage{acl}

\usepackage{times}
\usepackage{latexsym}

\usepackage[T1]{fontenc}

\usepackage[utf8]{inputenc}

\usepackage{microtype}

\usepackage{soul}
\usepackage{multicol}
\usepackage{multirow}
\usepackage{booktabs}
\usepackage{graphicx}
\usepackage{tabularx}
\usepackage[normalem]{ulem}
\useunder{\uline}{\ul}{}
\usepackage{amssymb}
\usepackage{tablefootnote} 
\usepackage[symbol]{footmisc}

\usepackage{subcaption}
\usepackage[inline]{enumitem}

\title{SueNes: A Weakly Supervised Approach to Evaluating Single-Document Summarization via Negative Sampling}

\author{Forrest Sheng Bao \and Ge Luo \and Hebi Li\\
  Iowa State University, Ames, IA, USA \\
  \texttt{forrest.bao@gmail.com}, 
  \texttt{\{gluo,hebi\}@iastate.edu} \\\AND
  Minghui Qiu \\
  Alibaba Group \\ Hangzhou, China \\
  \texttt{minghui.qmh@alibaba-inc.com}\\\And
  Yinfei Yang \\
  1600 Amphitheater Parkway \\ Mountain View, CA, USA \\
  \texttt{yangyin7@gmail.com} \\\AND
  Youbiao He \\
  Iowa State University \\ Ames, IA, USA \\
  \texttt{yh54@iastate.edu} \\\And
    Cen Chen \\
  East China Normal University\\ Shanghai, China\\
  \texttt{cecilia.cenchen@gmail.com} \\
}

\begin{document}
\maketitle
\begin{abstract}
Canonical automatic summary evaluation metrics, such as ROUGE, 
focus on lexical similarity which cannot well capture semantics nor linguistic quality
and require a reference summary which is costly to obtain.
Recently, 
there have been a growing number of efforts 
to alleviate either or both of the two drawbacks. 
In this paper, we present a proof-of-concept study to a weakly supervised summary evaluation approach without the presence of reference summaries.
Massive data in existing summarization datasets are transformed for training by pairing documents with corrupted reference summaries.
In cross-domain tests, our strategy outperforms baselines with promising improvements, and show a great advantage in gauging linguistic qualities over all metrics. 

\end{abstract}

\section{Introduction}

In natural language processing, the problem of summarization studies generating a summary from a source document which is longer than the summary. 
De facto metrics to judge a generated summary include  ROUGE~\cite{2004-Workshop-Lin-Rouge}, BLEU~\cite{2002-ACL-Papineni-Bleu}, and METEOR~\cite{2005-Workshop-Banerjee-METEOR}. 
Previous work~\cite{2015-EMNLP-Ng-Better,2008-ACL-Liu-Correlation,2016-EMNLP-Liu-Not,2018-ACL-Shang-Unsupervised} agrees on two major drawbacks of them: 1) they favor lexical similarity, falling short on semantic similarity or linguistic quality, and 2) they require a reference summary which is often expensive to obtain~\cite{2018-NAACL-Zopf-Estimating}.


Initially, the first drawback is partially alleviated by replacing lexicons with their word embeddings~\cite{2015-EMNLP-Ng-Better, 
2017-Workshop-Ellouze-Machine,
2018-Unknown-Ruseti-Scoring, 2019-Preprint-Xia-Automatic}. 
After the birth of transformers~\cite{2017-NIPS-Vaswani-Attention}, this effort has expanded to sentence or document level, including reference-based~\cite{bert-score,zhao-etal-2019-moverscore}, and reference-free ones~\cite{vasilyev-etal-2020-fill,scialom-etal-2019-answers,gao-etal-2020-supert}. 
The main difference between the two groups is whether a reference summary is needed when evaluating a machine-generated summary. 


The two groups have complementary pros and cons. 
Reference-based metrics have a better performance, but they are impractical when summarization is used industrially, such as in customer support~\cite{sum_customer}, team conversation~\cite{sum_groupchat}, and bug reporting~\cite{sum_bugreport}, where
it is too costly to manually craft an equally massive amount of reference summaries. 
In contrast, without human written reference summaries, reference-free approaches generally perform poorer. 
Modern transformer-based reference-free approaches  often rely on non-summarization tasks, such as QA~\cite{vasilyev-etal-2020-fill, scialom-etal-2019-answers}.
Such fact-focused approach makes them excel on content/fact aspects (still worse than reference-based ones) but not on linguistic ones. 
The non-summarization tasks also introduce noises. 


Therefore, in this paper, as a proof of concept, we explore a hybrid or middle approach to combine the best of both worlds.
Using document-summary pairs in existing summarization datasets, 
our weakly supervised approach mutates\footnote{We avoid the term ``augment'' here because ``augment'' means making something better but what we are doing here is corrupting reference summaries.} reference summaries and pair them with documents to form training data and then use the trained model to evaluate unseen summaries in the presence of documents without corresponding reference summaries.
In this way, we make use of human written summaries, which are very precious, in training, but we do not need them in summary evaluation. 
We call our approach \textbf{SueNes}
, which stands for ``\textbf{Su}mmary \textbf{e}valuation by \textbf{Ne}gative \textbf{s}ampling.'' 

The quality of a summary is usually evaluated on two facets: content/fact aspects and linguistic qualities. 
Experiments later will show that a value of our approach is that we can use the same model architecture to build models that excel on different tasks by 
feeding training data from the same source but mutated in different strategies.
For example, deleting words is the best for linguistic qualities while deleting sentences  is the best for content/fact coverage. 

Our approach is empirically compared against an array of existing metrics on three human summary evaluation datasets.
Despite being training-based, 
our approach exhibits consistent results across various training domains which are all different from the test domain.
It outperforms reference-free baselines with promising improvements on content/fact aspects, and further outperforms all existing metrics in gauging linguistic qualities. 

In summary, our contributions or merits are:
\begin{itemize}
  \setlength\itemsep{0em}
 \item a simple but effective approach to reference-free\footnote{The definition of ``reference-free'' is that reference summaries are not needed in the evaluation stage. } summary quality assessment, 
 \item negative sampling for preparing training data from the unlabeled,
 \item one task/framework for multi-aspect judging,
 \item extensive cross-domain experiments to validate the effectiveness and domain robustness of our approach. 
%
\end{itemize}

    We hope our study can inspire more research into hybridizing reference-free and reference-based summary evaluation. 
Our  code is at \url{https://github.com/forrestbao/SueNes/} 

\section{The Approach}

\subsection{Model Architecture}
\label{sec:model_architecture}
A reference-free single-document summary quality assessor can be formulated as a regression function $f(d,s)\in[0,1]$ of an input document $d= [t_1, t_2, \cdots]$,
and a machine-generated summary $s=[t'_1, t'_2, \cdots]$, 
where $t_i$'s and $t'_i$'s are text tokens. 
As a proof of concept, we explore an extremely lean implementation of $f$: first 
$d$ and $s$
are jointly transformed into a vector representation $\mathbf{e}=g(d,s)$, and then it is mapped to a summary quality score via a fully-connected layer, i.e.,  $f(d,s) = \sigma (\mathbb{W} \mathbf{e})$. 

The function $g$ can be implemented in the 
BERT~\cite{Devlin2018} style
with an input sequence
\texttt{[[CLS]}, $t_1$, $t_2$, $\cdots$, \texttt{[SEP]}, $t'_1$, $t'_2$, $\cdots$, \texttt{[SEP]]}.
The output on the \texttt{[CLS]} token 
is a joint representation of both the document $d$ and the summary $s$.

While the human evaluation to a summary may cover multiple aspects, such as content/fact coverage and linguistics, a model of us will only yield one number. 
But by using 
 different data mutation strategies, we can get models (different $f$'s) adept at different aspects of a summary.


\begin{figure*}[!htbp]
    \centering
    \includegraphics[width=0.9\textwidth]{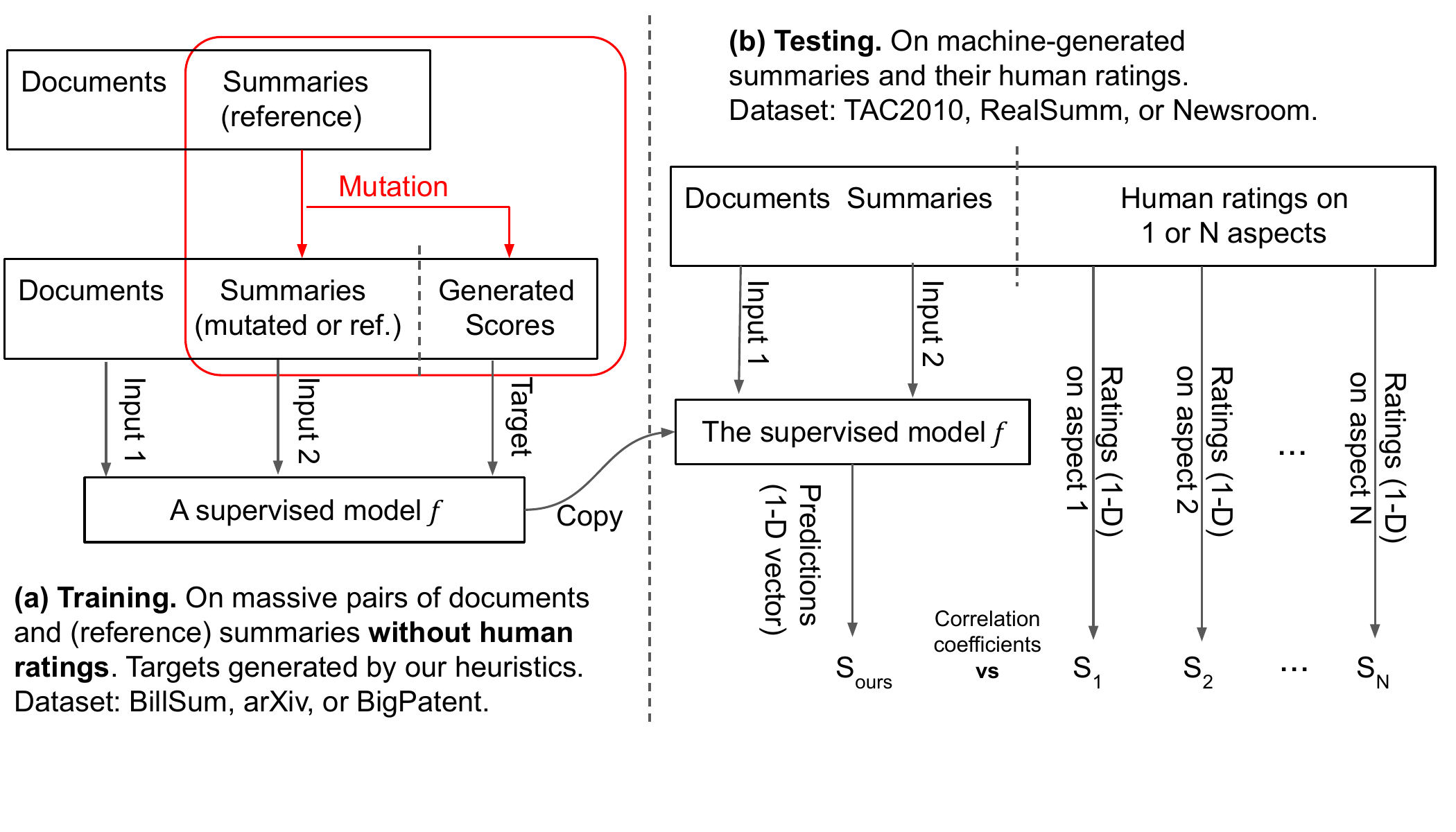}
    \caption{The weakly supervised training approach in this paper and the test of a trained model.}
    \label{fig:workflow}
\end{figure*}

\subsection{Negative Sample Generation}

\label{sec:data_aug}



It is impractical to train $f$ with existing summarization datasets, such as CNN/DailyMail~\cite{2015-NIPS-Hermann-Teaching,2016-CoNLL-Nallapati-Abstractive}, because they contain only high-quality, reference-class summaries written manually and thus are all of label 1. 
Some summary evaluation datasets, such as RealSumm~\cite{realsumm}, Newsroom~\cite{newsroom}, and TAC2010~\cite{TAC2010},
do contain human ratings to system-generated summaries of various qualities. 
But they are too small, containing no more than 100 news articles or article groups each. 
Therefore, training against human ratings or in a supervised approach is impractical. 

\begin{figure}[!htbp]
    \centering
    \includegraphics[width=.9\columnwidth]{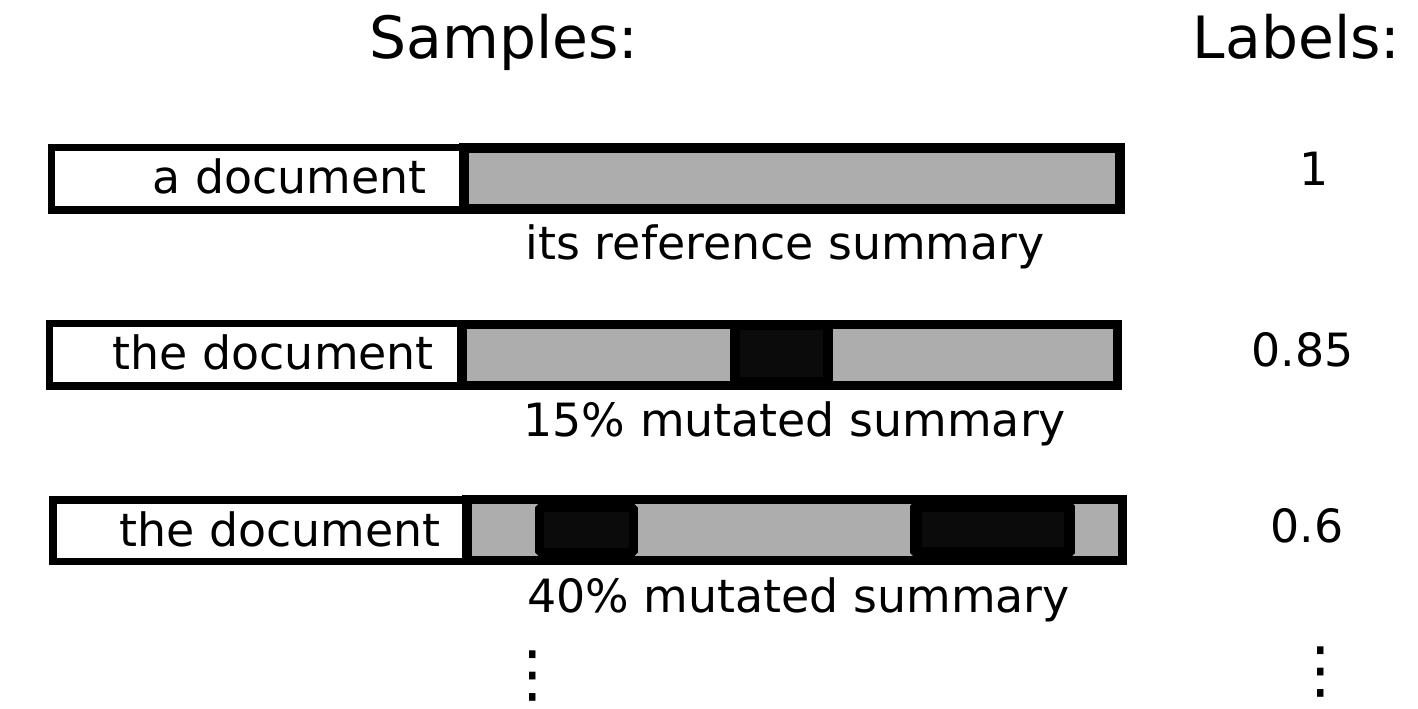}
    \caption{Training sample generation by mutation. Mutated text in dark blocks \rule{1.7em}{0.7em} while intact text in gray blocks \textcolor{gray}{\rule{1.6em}{.7em}}. Sizes are out of scale. }
    \label{fig:mutate}
\end{figure}

To work around, we propose a weakly supervised solution as depicted in Figure~\ref{fig:workflow}(a).
Existing summarization datasets contain many document-summary pairs. 
For each pair $\langle d,s \rangle $, the reference summary $s$
is mutated into $K$ new summaries of different extents $s_1, s_2, \cdots, s_K$, which are then paired with the document to form new pairs $$\langle d, s_1 \rangle,  \langle d, s_2 \rangle, \cdots, \langle d, s_K \rangle, $$ which are finally assigned targets to form the training data $$(\langle d, s_1 \rangle, y_1),  (\langle d, s_2 \rangle, y_2), \cdots, (\langle d, s_K \rangle, y_K).$$
As illustrated in Figure~\ref{fig:mutate},  the training target $y_{k\in[1..K]}$ is the percentage of intact content. 
For example, if 30\% of tokens in a mutated summary are not original,
then the label is 0.7. 
In addition, the original document-summary pair $\langle d, s \rangle$ is also used in training with a target of 1. 

Mutations can happen at the token or sentence level, 
where tokens or sentences are randomly selected for mutation. 
A selected token or sentence is mutated in one of the three methods: 
\begin{enumerate}
    \item \textbf{inserting} a token/sentence from other summaries behind it, 
    \item \textbf{deleting} it, or
    \item \textbf{replacing} it with a token/sentence from other summaries.
\end{enumerate}
We do not mix different mutation levels nor mix different mutation methods when preparing the training data. 
Instead, our experiments study one combination of a mutation level and a mutation method, denoted as a mutation strategy,  each time.

\section{Experiments}
\label{sec:exp}

\subsection{Test data}
\label{sec:data_basic}

The ground truth of a summary's quality is human ratings to it.
A model trained (Fig.~\ref{fig:workflow}(a)) is tested (Fig.~\ref{fig:workflow}(b)) against human ratings. 
Three test datasets are chosen below. 
Due to the limited number and sizes of human evaluation datasets, they are all in the news domain. 
The human evaluation protocols can be found in their respective references. 


\textbf{TAC2010}~\cite{TAC2010} is a multi-document (ten-document) summarization task reporting both factual and linguistic aspects. 
We use $\sum_{i\in [1..10]} f(d_i,s)$ to approximate the score of the summary $s$ 
composed from ten documents $d_1$ to $d_{10}$. 
We only use Set A of TAC2010 because Set B is not for regular summarization. 

\textbf{Newsroom}~\cite{newsroom} also covers both factual (in INFormativeness and RELevance) and linguistic (in COHerence and FLUency) aspects. For human ratings, three human annotators rate one pair of a document and machine-generated summary. The mean of their ratings on each aspect is used in our experiments.  

\textbf{RealSumm}~\cite{realsumm} focuses on only factual coverage. It covers 14 abstractive and 11 extractive summarizers published after 2018 and conducts human evaluation on the two groups separately. 

Note that we do not and cannot train a model against the labels in a test set, as mentioned in \S~\ref{sec:model_architecture}. 
If a test set rates on multiple aspects, we do not train one model for each aspect. 
Nor do we train models for individual or a collection of test sets. 
We compute correlation between the predictions from our model and human ratings on each aspect of each test set.

\subsection{Training data}
\label{sec:training_data}
Three widely used summarization datasets from three different domains are chosen for training:
\textbf{Billsum}~\cite{billsum},  \textbf{Scientific-Papers/arXiv}~\cite{scientific_papers}, and  \textbf{Big-Patent}~\cite{bigpatent}.
Datasets from the news domain are avoided on purpose because the test data is in the news domain. 
This cross-domain setting allows us to
examine whether a model is prone to domain differences. 
For each reference summary, $K=5$ mutated summaries are generated. 
The percentage of intact content is measured by the number of tokens and the number of characters for token-level and sentence-level mutations, respectively.

\subsection{Baselines and upper bounds}


To fairly compare, four recent metrics: \textbf{BLANC}~\cite{vasilyev-etal-2020-fill}, \textbf{SummaQA}~\cite{scialom-etal-2019-answers}, \textbf{SUPERT}~\cite{gao-etal-2020-supert} 
and \textbf{LS-Score}~\cite{LS_Score}
, are used as
baselines because like our approach, they do not need a reference summary to judge a machine-generated summary, i.e., reference-free. 

Human crafted reference summaries give reference-based metrics advantages.
The results of reference-based metrics are included 
as soft upper bounds:  
     \textbf{ROUGE-1, ROUGE-2 and ROUGE-L}~\cite{2004-Workshop-Lin-Rouge},
     \textbf{MoverScore}~\cite{zhao-etal-2019-moverscore},
     \textbf{BertScore}~\cite{bert-score},
     \textbf{BLEU}~\cite{2002-ACL-Papineni-Bleu},
     \textbf{METEOR}~\cite{2005-Workshop-Banerjee-METEOR}, and
    $\textbf{S}^3$~\cite{peyrard-etal-2017-learning}.

\subsection{Settings}
Because the baselines use BERT, we use BERT as well for a fair comparison. 
Specifically, BERT-base uncased (L=12, H=768) is fine-tuned, with a learning rate of 1e-5, three epochs, and a batch size of 14. 
The input sequence is limited to 512 tokens using the round robin trimmer. 
The training loss is MSE as this problem is regression. 

\subsection{Results}
\label{sec:results}

We use the summary-level~\cite{peyrard-etal-2017-learning} meta-evaluation strategy to report an approach's average correlation with human ratings. 
Summary evaluation usually covers two types of aspects, contents/facts and linguistics. 
They are reported separately in Tables~\ref{tab:fact} and ~\ref{tab:ling}.
Due to space limit, only the best mutation strategy is reported for each aspect group. 

\textbf{On content/fact aspects}, 
the best mutation strategy is sentence deletion and our best models outperform baselines on all test datasets. 
Our approach makes the most improvement over baselines on RealSumm, a dataset much bigger than Newsroom and more modern than TAC2010, and the least improvement on TAC2010, the oldest dataset. 


\begin{table}[!htbp]
\setlength{\tabcolsep}{0.2em} 
\centering
\scriptsize
\caption{Spearman's correlation on \textbf{content/fact} aspects.
{\tiny Superscripts are ranks per aspect. Abs. and Ext. are two summarizer groups in RealSumm. }}\label{tab:fact}
\vskip -1em
\begin{tabular}{lcc|cc|cc}
\toprule
{}                                                                            & {} & TAC2010       & \multicolumn{2}{c|}{Newsroom}            & \multicolumn{2}{c}{RealSumm}  \\
{}                                                                            & {} & Pyramid       & {INF} & REL           & Abs.          & Ext.
\\
\midrule
\multirow{4}{1.8cm}{Our approach (\textit{mutated in sentence deletion})} & \textit{Trained on}:  &  &&&&\\
 & Billsum              & \textbf{0.49}$^1$ & {\ul 0.70}$^2$              & $\underline{0.61}^3$    & 0.26          & 0.01          \\
                                                                                                & arXiv                & 0.41          & 0.69                    & 0.59          & \textbf{0.34}$^1$ & {\ul 0.12}$^2$    \\
                                                                                                & BigPatent             & 0.42          & \textbf{0.75}$^1$           & \textbf{0.65}$^1$ & {\ul 0.33}$^2$    & \textbf{0.13}$^1$ \\
\midrule
{\multirow{4}{*}{Baselines}}                                                           & BLANC-tune           & {\ul 0.43}$^3$    & 0.69                    & {\ul 0.61}$^2$    & {\ul 0.31}$^3$    & {\ul 0.11}$^3$    \\
{}                                                                            & SummaQA-F1           & 0.30          & 0.57                    & 0.52          & 0.22          & 0.08          \\
{}                                                                            & SummaQA-CFD          & 0.29          & 0.54                    & 0.44          & 0.24          & 0.05          \\
{}                                                                            & SUPERT               & {\ul 0.48}$^2$    & {\ul 0.69}$^3$              & 0.60          & 0.25          & 0.07          \\
  & LS-Score * & N/A & 0.70 & 0.64 & N/A & N/A \\

\midrule
\multirow{9}{*}{Upper bounds}                                                                   & R-1                  & 0.56          & 0.32                    & 0.28          & 0.63          & 0.22          \\
                                                                                                & R-2                  & 0.64          & 0.15                    & 0.13          & 0.56          & 0.22          \\
                                                                                                & R-L                  & 0.50          & 0.30                    & 0.26          & 0.60          & 0.21          \\
                                                                                                & MoverScore           & 0.72          & 0.22                    & 0.22          & 0.50          & 0.19          \\
                                                                                                & BertScore            & 0.68          & 0.32                    & 0.28          & 0.57          & 0.19          \\
                                                                                                & BLEU                 & 0.60          & -0.08                   & -0.01         & 0.30          & 0.16          \\
                                                                                                & METEOR               & 0.67          & 0.24                    & 0.24          & 0.63          & 0.25          \\
                                                                                                & S3\_pyr              & 0.73          & 0.27                    & 0.25          & 0.64          & 0.24          \\
                                                                                                & S3\_resp             & 0.73          & 0.25                    & 0.22          & 0.63          & 0.24          \\
\midrule
\multicolumn{2}{c}{Our best over baseline best (\%)}                                                                   & 2.71          & 8.67                    & 6.40          & 9.72          & 16.42         \\
\multicolumn{2}{c}{Our average absolute deviation (\%)}                                                                            & 3.32          & 2.57                    & 2.21          & 3.45          & 5.28          \\
\bottomrule
\end{tabular}

\end{table}


\textbf{On linguistic aspects}, the best mutation strategy is word deletion. 
Here, even our worst model cannot be outperformed by any baseline nor upper bound. 
As mentioned earlier, canonical metrics are lexical-based while modern reference-based and reference-free approaches focus on facts. 
Through mutating reference summaries, our approach can create summaries of different linguistic qualities. 
Although our approach makes big improvements over baselines on TAC2010 and Newsroom's FLUency, its edge is smaller on Newsroom's COHerence. 
A sentence-level scrambling mutation may improve our approach's performance on COHerence in the future. 

\begin{table}[!htbp]
\centering
\scriptsize
\caption{Spearman's correlation on \textbf{linguistic} aspects. {\tiny Superscripts are ranks in each aspect/column.} }\label{tab:ling}
\vskip -1em 
\begin{tabular}{ccc|cc}
\toprule
{}                       & {} & TAC2010       & \multicolumn{2}{c}{Newsroom}  \\
{}                       & {} & Ling.         & COH           & FLU           \\
\midrule
\multirow{4}{1.5cm}{Our approach (\textit{mutated in word deletion})} &  \textit{Trained on:}  & & &  \\
& Billsum              & \textbf{0.46}$^1$ & {\ul 0.65}$^2$    & {\ul 0.65}$^2$    \\
                                           & arXiv                & {\ul 0.38}$^3$    & \textbf{0.67}$^1$ & \textbf{0.67}$^1$ \\
                                           & BigPatent            & {\ul 0.43}$^2$    & {\ul 0.62}$^3$    & {\ul 0.63}$^3$    \\
\midrule
\multirow{4}{*}{Baselines}                          & BLANC-tune           & 0.29          & 0.59          & 0.53          \\
                                           & SummaQA-F1           & 0.24          & 0.49          & 0.47          \\
                                           & SummaQA-CFD          & 0.15          & 0.42          & 0.37          \\
                                           & SUPERT               & 0.32          & {\ul 0.62}$^2$          & 0.54          \\
                                           
  & LS-Score \tablefootnote[1]{\tiny LS\_Score results are only for Newsroom, which are copied from its paper, as we cannot run their code on other datasets after trying really hard. Several other researchers reported the same issue at \url{https://github.com/whl97/LS-Score/issues}. It is further excluded from the ranking because it is trained on the same domain as the test domain whereas all other baselines and our models are not. } & N/A & 0.63 & 0.59 \\                                           
\midrule
\multirow{9}{*}{Upper bounds}              & R-1                  & 0.26          & 0.23          & 0.22          \\
                                           & R-2                  & 0.35          & 0.09          & 0.10          \\
                                           & R-L                  & 0.18          & 0.21          & 0.20          \\
                                           & MoverScore           & 0.35          & 0.17          & 0.14          \\
                                           & BertScore            & 0.36          & 0.27          & 0.24          \\
                                           & BLEU                 & 0.35          & -0.06         & -0.04         \\
                                           & METEOR               & 0.34          & 0.17          & 0.17          \\
                                           & S3\_pyr              & 0.36          & 0.19          & 0.18          \\
                                           & S3\_resp             & 0.36          & 0.17          & 0.17          \\
\midrule
\multicolumn{2}{c}{Our best over baseline best (\%)}              & 41.92         & 8.41          & 25.02         \\
\multicolumn{2}{c}{Our average absolute deviation (\%)}                       & 2.72          & 1.71          & 1.74         \\
\bottomrule
\end{tabular}
\end{table}

\subsection{Discussions} 

\textbf{What is the best mutation? }
Across datasets, deletion-based mutations are most effective. 
The two kinds of deletions happen to  be complementarily effective for two aspect groups: sentence deletion for content/fact aspects vs. word deletion for linguistic aspects. 
This is an advantage of our approach that \textit{under a uniformed framework, different summary quality aspects can be gauged by designing different mutation options}. 

The complementariness of sentence deletion and word deletion can be well explained as that removing a sentence from a reference summary reduces a great amount of key information while removing a word from a sentence changes it syntactically. 
We found that word-level mutations are less useful for content/fact aspects, probably because of the inertia of the context after words are altered. 

%

\textbf{Which training domain/dataset should I use?}
Due to the composition of summarizers and the limited  data size in human evaluation, it is very hard to get a consistent ranking of metrics on different datasets~\cite{realsumm}. 
For example, in Table~\ref{tab:fact}, Billsumm outperforms all baselines and its peers on TAC2010  but not the case on Newsroom and RealSumm. 

Still, the impact of training domain seems manageable. The average absolute deviations across the training datasets/domains are given at the bottom of Tables~\ref{tab:fact} and \ref{tab:ling}.
They mostly below 3.5\%. 
A qualitative analysis shows that the variation seems more due to the characteristics of the text than the domain. Legislative bills (Billsum) have lots of short, hierarchical clauses and thus differ from common English greatly. 
Scientific papers have many equations and cross-references. 
There are also many occurrences of \LaTeX~or MathML in the dataset arXiv. 
On top of that, all our experiments use different training and test domains. 
Hence we would say that the impact of domain variation is very small.

\section{Conclusion}
In this paper, we propose a weakly supervised approach to summary quality evaluation.
A few mutation methods are introduced to make use of the massive, precious human written summaries in summarization datasets.
In cross-domain experiments, our approach achieves better performance than baselines, especially on linguistic aspects. 
We hope this proof-of-concept study can inspire more reference-free summary evaluation. 

\section*{Acknowledgments}
Bao, Luo, Li, and He's work in this paper is partially supported by National Science Foundation (NSF) grants No. MCB-1821828 and No. CNS-1817089. 
The authors would also like to thank reviewers who have given precious feedback on improving this work.

\bibliography{aaai22}
\bibliographystyle{acl_natbib}

\newpage

\appendix



\section{Dataset statistics}
For test sets:
\begin{itemize}
    \item \textbf{TAC2010 Guided Summarization Task Set A} consists of 46 topics, each of which is associated with a set of 10 documents. We evaluate the metrics over summaries generated by 43 systems.
    \item \textbf{Newsroom} contains human-rated summaries generated by 7 systems for 60 documents.
    \item \textbf{RealSumm} sampled 100 documents from the CNN/DailyMail test set, and collected human ratings for summaries generated by 11 extrative systems and 14 abstractive systems.
\end{itemize}

For training sets, the numbers of pairs of documents and reference summaries in the train split are:
\begin{itemize}
    \item \textbf{Billsum}: 18,949
    \item \textbf{Scientific papers/arXiv}: 203,037
    \item \textbf{Big-Patent}:   1,207,222
\end{itemize}
For each dataset, we use the entire (except for BigPatent, 10\% due to its huge size) \texttt{train} split in Google Tensorflow Datasets for training. 

\section{Computational environment and cost}
All experiments were carried out on one RTX3090 GPU installed on a desktop computer. The training takes about a week for all three training datasets.

\section{Another type of mutation}
In addition to the three mutation methods mentioned already, we have another method called crosspairing. 

\begin{figure}[!htbp]
  \includegraphics[width=\columnwidth]{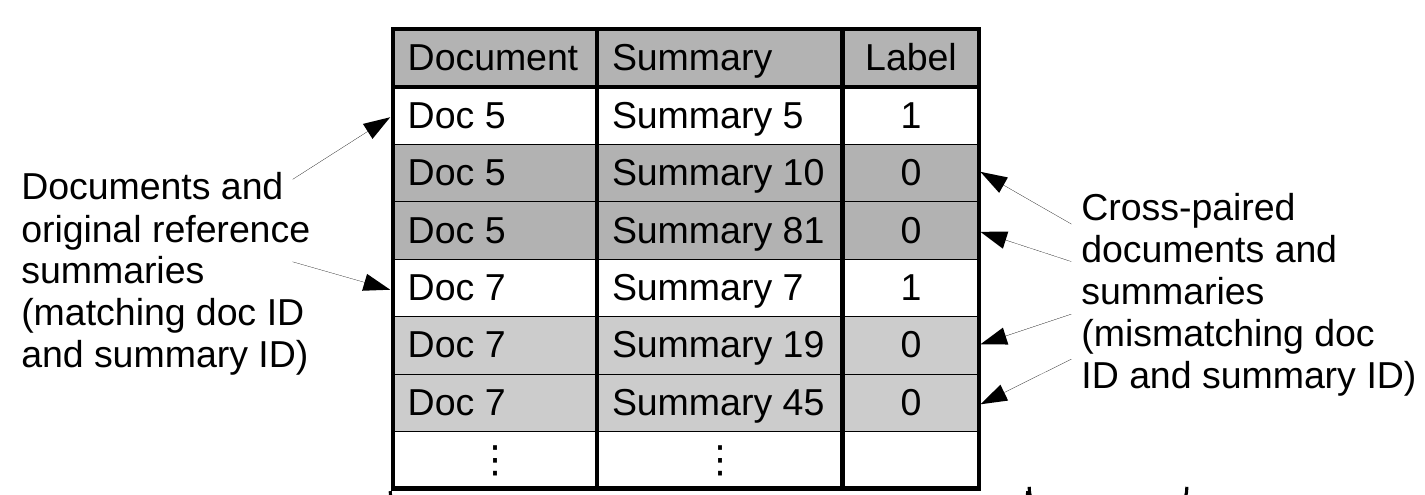}
  \caption{Training sample generation via cross pairing.}
  \label{fig:cross-pairing}
\end{figure}

Illustrated in Figure~\ref{fig:cross-pairing}, 
it is inspired by 
the next-sentence prediction (NSP) task in original BERT training. 
Given a document and its reference summary, we create negative data by pairing the document with reference summaries of other documents.
We assign the label 0 to a mismatching document-summary pair, and the label 1 to any original pair of a document and its reference summary.

\section{Complete empirical results}

Due to space limit, we were only able to present the result of the best mutation method in \S~\ref{sec:results}. 
Full results are given in Tables~\ref{tab:fact_rho},~\ref{tab:ling_rho},~\ref{tab:fact_r}, and~\ref{tab:ling_r}. 
Pearson's for LS-Score is unable to be produced due to reasons explained in the footnote on page 4. 

\begin{table}[!htb]
\setlength{\tabcolsep}{0.1em} 
\centering
\scriptsize
\caption{Full results for \textbf{Spearman}'s correlation on \textbf{content/fact} aspects.}\label{tab:fact_rho}
\begin{tabular}{cccc|cc|cc}
\toprule
{}                                 &    \multirow{2}{*}{Mutation}                                   & {Training} & {TAC2010} & \multicolumn{2}{c|}{Newsroom}                         & \multicolumn{2}{c}{RealSumm}                            \\
{}                                 &  & set         & {Pyramid} & INF                       & REL                      & Abs                        & Ext                        \\
\midrule
\multirow{18}{1.5cm}{Our approach}                       & \multirow{3}{1.5cm}{crosspair}            & Billsum              & 0.38                        & 0.50                      & 0.49                     & -0.06                      & -0.05                      \\
                                                     &                                       & ArXiv                & 0.37                        & 0.57                      & 0.55                     & -0.06                      & -0.08                      \\
                                                     &                                       & BigPatent            & 0.33                        & 0.56                      & 0.57                     & -0.06                      & -0.05                      \\
\cmidrule{2-8}
                                                     & \multirow{3}{1.5cm}{sentence-replace}     & Billsum              & 0.44                        & 0.47                      & 0.42                     & 0.04                       & -0.08                      \\
                                                     &                                       & ArXiv                & 0.35                        & 0.55                      & 0.49                     & 0.19                       & 0.03                       \\
                                                     &                                       & BigPatent            & 0.39                        & 0.49                      & 0.46                     & -0.08                      & -0.04                      \\
\cmidrule{2-8}
                                                     & \multirow{3}{1.5cm}{word-insert}             & Billsum              & 0.21                        & 0.60                      & 0.56                     & 0.06                       & -0.01                      \\
                                                     &                                       & ArXiv                & 0.10                        & 0.66                      & 0.58                     & 0.20                       & -0.01                      \\
                                                     &                                       & BigPatent            & 0.20                        & 0.63                      & 0.59                     & 0.14                       & -0.02                      \\
\cmidrule{2-8}
                                                     & \multirow{3}{1.5cm}{word-delete}          & Billsum              & 0.27                        & 0.64                      & 0.61                     & 0.12                       & 0.02                       \\
                                                     &                                       & ArXiv                & 0.23                        & 0.62                      & 0.59                     & 0.17                       & 0.01                       \\
                                                     &                                       & BigPatent            & 0.28                        & 0.59                      & 0.60                     & 0.10                       & 0.01                       \\
\cmidrule{2-8}
                                                     & \multirow{3}{1.5cm}{word-replace}         & Billsum              & 0.25                        & 0.66                      & 0.60                     & 0.10                       & -0.03                      \\
                                                     &                                       & ArXiv                & 0.08                        & 0.65                      & 0.57                     & 0.15                       & -0.02                      \\
                                                     &                                       & BigPatent            & 0.25                        & 0.63                      & 0.62                     & 0.07                       & -0.06                      \\
\cmidrule{2-8}
                                                     & \multirow{3}{1.5cm}{sentence-delete}      & Billsum              & 0.49                        & 0.70                      & 0.61                     & 0.26                       & 0.01                       \\
                                                     &                                       & ArXiv                & 0.41                        & 0.69                      & 0.59                     & 0.34                       & 0.12                       \\
                                                     &                                       & BigPatent            & 0.42                        & 0.75                      & 0.65                     & 0.33                       & 0.13                       \\
\midrule
\multirow{5}{*}{Baselines}                           & \multicolumn{2}{c}{BLANC-tune}                               & 0.43    & 0.69                      & 0.61                     & 0.31                       & 0.11                       \\
                                                     & \multicolumn{2}{c}{SummaQA-F1}                               & 0.30    & 0.57                      & 0.52                     & 0.22                       & 0.08                       \\
                                                     & \multicolumn{2}{c}{SummaQA-CFD}                              & 0.29    & 0.54                      & 0.44                     & 0.24                       & 0.05                       \\
                                                     & \multicolumn{2}{c}{SUPERT}                                   & 0.48    & 0.69                      & 0.60                     & 0.25                       & 0.07                       \\
                                                     & \multicolumn{2}{c}{LS-Score *}                              & N/A     & 0.70                      & 0.64                     & N/A                        & N/A                        \\
\midrule
\multirow{9}{*}{Upper bounds}                        & \multicolumn{2}{c}{R-1}                                      & 0.56    & 0.32                      & 0.28                     & 0.63                       & 0.22                       \\
                                                     & \multicolumn{2}{c}{R-2}                                      & 0.64   & 0.15                      & 0.13                     & 0.56                       & 0.22                       \\
                                                     & \multicolumn{2}{c}{R-L}                                      & 0.50    & 0.30                      & 0.26                     & 0.60                       & 0.21                       \\
                                                     & \multicolumn{2}{c}{MoverScore}                               & 0.72    & 0.22                      & 0.22                     & 0.50                       & 0.19                       \\
                                                     & \multicolumn{2}{c}{BertScore}                                & 0.68    & 0.32                      & 0.28                     & 0.57                       & 0.19                       \\
                                                     & \multicolumn{2}{c}{BLEU}                                     & 0.60    & -0.08                     & -0.01                    & 0.30                       & 0.16                       \\
                                                     & \multicolumn{2}{c}{METEOR}                                   & 0.67    & 0.24                      & 0.24                     & 0.63                       & 0.25                       \\
                                                     & \multicolumn{2}{c}{S3\_pyr}                                  & 0.73    & 0.27                      & 0.25                     & 0.64                       & 0.24                       \\
                                                     & \multicolumn{2}{c}{S3\_resp}                                 & 0.73    & 0.25                      & 0.22                     & 0.63                       & 0.24                       \\
\midrule
\multicolumn{3}{c}{Our best over baseline best (\%)}                                                                & -8.47                       & {-4.63} & {2.14} & {-35.93} & {-76.38} \\
\midrule
\multirow{6}{1.5cm}{Our average absolute deviation (\%)} & \multicolumn{2}{c}{crosspair}                                & 2.02                        & {2.75}  & {3.00} & {0.00}   & {1.02}   \\
                                                     & \multicolumn{2}{c}{sentence-delete}                          & 3.32                        & {2.57}  & {2.21} & {3.45}   & {5.28}   \\
                                                     & \multicolumn{2}{c}{sentence-replace}                         & 2.99                        & {3.34}  & {2.57} & {9.28}   & {3.92}   \\
                                                     & \multicolumn{2}{c}{word-insert}                                 & 4.64                        & {1.87}  & {1.03} & {5.01}   & {0.37}   \\
                                                     & \multicolumn{2}{c}{word-delete}                              & 1.96                        & {1.79}  & {0.82} & {2.55}   & {0.44}   \\
                                                     & \multicolumn{2}{c}{word-replace}                             & 7.59                        & {1.11}  & {1.73} & {2.60}   & {1.96}  \\
\bottomrule
\end{tabular}
\end{table}

\begin{table}[!htb]
\setlength{\tabcolsep}{0.2em} 
\centering
\scriptsize
\caption{Full results for \textbf{Spearman}'s correlation on \textbf{linguistic} aspects.}\label{tab:ling_rho}
\begin{tabular}{cccc|cc}
\toprule
                                                     &  \multirow{2}{*}{Mutation}                                 &  Training            & TAC2010    & \multicolumn{2}{c}{Newsroom} \\
                                                     &                & set & Linguistic & COH           & FLU          \\
\midrule
\multirow{18}{1.5cm}{Our approach}                       & \multirow{3}{1.5cm}{crosspair}        & Billsum      & 0.29       & 0.43          & 0.39         \\
                                                     &                                   & ArXiv        & 0.28       & 0.48          & 0.42         \\
                                                     &                                   & BigPatent    & 0.28       & 0.48          & 0.42         \\
\cmidrule{2-6}
                                                     & \multirow{3}{1.5cm}{sentence-delete}  & Billsum      & 0.33       & 0.59          & 0.53         \\
                                                     &                                   & ArXiv        & 0.32       & 0.53          & 0.46         \\
                                                     &                                   & BigPatent    & 0.30       & 0.62          & 0.54         \\
\cmidrule{2-6}
                                                     & \multirow{3}{1.5cm}{sentence-replace} & Billsum      & 0.39       & 0.45          & 0.42         \\
                                                     &                                   & ArXiv        & 0.27       & 0.50          & 0.43         \\
                                                     &                                   & BigPatent    & 0.38       & 0.41          & 0.31         \\
\cmidrule{2-6}
                                                     & \multirow{3}{1.5cm}{word-insert}         & Billsum      & 0.31       & 0.55          & 0.53         \\
                                                     &                                   & ArXiv        & 0.16       & 0.55          & 0.48         \\
                                                     &                                   & BigPatent    & 0.19       & 0.51          & 0.48         \\
 \cmidrule{2-6}
                                                     & \multirow{3}{1.5cm}{word-replace}     & Billsum      & 0.33       & 0.60          & 0.57         \\
                                                     &                                   & ArXiv        & 0.07       & 0.54          & 0.49         \\
                                                     &                                   & BigPatent    & 0.24       & 0.54          & 0.46         \\
\cmidrule{2-6}
                                                     & \multirow{3}{1.5cm}{word-delete}      & Billsum      & 0.46       & 0.65          & 0.65         \\
                                                     &                                   & ArXiv        & 0.38       & 0.67          & 0.67         \\
                                                     &                                   & BigPatent    & 0.43       & 0.62          & 0.63         \\
\midrule
\multirow{5}{*}{Baselines}                           & \multicolumn{2}{c}{BLANC-tune}                   & 0.29       & 0.59          & 0.53         \\
                                                     & \multicolumn{2}{c}{SummaQA-F1}                   & 0.24       & 0.49          & 0.47         \\
                                                     & \multicolumn{2}{c}{SummaQA-CFD}                  & 0.15       & 0.42          & 0.37         \\
                                                     & \multicolumn{2}{c}{SUPERT}                       & 0.32       & 0.62          & 0.54         \\
                                                     & \multicolumn{2}{c}{LS-Score *}                  & N/A        & 0.63          & 0.59         \\
\midrule
\multirow{9}{*}{Upper bounds}                        & \multicolumn{2}{c}{R-1}                          & 0.26       & 0.23          & 0.22         \\
                                                     & \multicolumn{2}{c}{R-2}                          & 0.35       & 0.09          & 0.10         \\
                                                     & \multicolumn{2}{c}{R-L}                          & 0.18       & 0.21          & 0.20         \\
                                                     & \multicolumn{2}{c}{MoverScore}                   & 0.35       & 0.17          & 0.14         \\
                                                     & \multicolumn{2}{c}{BertScore}                    & 0.36       & 0.27          & 0.24         \\
                                                     & \multicolumn{2}{c}{BLEU}                         & 0.35       & -0.06         & -0.04        \\
                                                     & \multicolumn{2}{c}{METEOR}                       & 0.34       & 0.17          & 0.17         \\
                                                     & \multicolumn{2}{c}{S3\_pyr}                      & 0.36       & 0.19          & 0.18         \\
                                                     & \multicolumn{2}{c}{S3\_resp}                     & 0.36       & 0.17          & 0.17         \\
\midrule
\multicolumn{3}{c}{Our best over baseline best (\%)}                                                    & 19.17      & -0.28         & 5.49         \\
\midrule
\multirow{6}{1.5cm}{Our average absolute deviation (\%)} & \multicolumn{2}{c}{crosspair}                    & 0.29       & 2.00          & 1.50         \\
                                                     & \multicolumn{2}{c}{sentence-delete}              & 1.15       & 3.10          & 3.17         \\
                                                     & \multicolumn{2}{c}{sentence-replace}             & 4.97       & 3.05          & 5.05         \\
                                                     & \multicolumn{2}{c}{word-insert}                     & 6.01       & 1.62          & 2.38         \\
                                                     & \multicolumn{2}{c}{word-delete}                  & 2.72       & 1.71          & 1.74         \\
                                                     & \multicolumn{2}{c}{word-replace}                 & 9.28       & 2.56          & 4.23        \\
 \bottomrule
\end{tabular}
\end{table}

\begin{table}[!htb]
\setlength{\tabcolsep}{0.1em} 
\centering
\scriptsize
\caption{Full results for \textbf{Pearson}'s correlation on \textbf{content/fact} aspects.}\label{tab:fact_r}
\begin{tabular}{cccc|cc|cc}
\toprule
                                                     & \multirow{2}{*}{Mutation}                                  & Training             & TAC2010 & \multicolumn{2}{c|}{Newsroom} & \multicolumn{2}{c}{RealSumm} \\
                                                     &                  & set & Pyramid & INF           & REL          & Abs           & Ext          \\
\midrule
\multirow{18}{1.5cm}{Our approach}                       & \multirow{3}{1.5cm}{crosspair}        & Billsum      & 0.44    & 0.63          & 0.66         & -0.07         & -0.05        \\
                                                     &                                   & ArXiv        & 0.45    & 0.62          & 0.65         & -0.07         & -0.07        \\
                                                     &                                   & BigPatent    & 0.39    & 0.63          & 0.68         & -0.07         & -0.05        \\
\cmidrule{2-8}
                                                     & \multirow{3}{1.5cm}{sentence-replace} & Billsum      & 0.48    & 0.64          & 0.67         & 0.04          & -0.09        \\
                                                     &                                   & ArXiv        & 0.24    & 0.56          & 0.58         & 0.07          & 0.05         \\
                                                     &                                   & BigPatent    & 0.41    & 0.59          & 0.61         & -0.07         & -0.04        \\
\cmidrule{2-8}
                                                     & \multirow{3}{1.5cm}{word-insert}         & Billsum      & 0.34    & 0.70          & 0.72         & 0.08          & 0.00         \\
                                                     &                                   & ArXiv        & 0.30    & 0.67          & 0.69         & 0.19          & -0.01        \\
                                                     &                                   & BigPatent    & 0.26    & 0.64          & 0.68         & 0.14          & -0.02        \\
\cmidrule{2-8}
                                                     & \multirow{3}{1.5cm}{word-delete}      & Billsum      & 0.39    & 0.76          & 0.78         & 0.12          & 0.05         \\
                                                     &                                   & ArXiv        & 0.39    & 0.68          & 0.70         & 0.18          & 0.03         \\
                                                     &                                   & BigPatent    & 0.38    & 0.71          & 0.74         & 0.13          & 0.01         \\
\cmidrule{2-8}
                                                     & \multirow{3}{1.5cm}{word-replace}     & Billsum      & 0.35    & 0.72          & 0.76         & 0.09          & -0.04        \\
                                                     &                                   & ArXiv        & 0.29    & 0.67          & 0.70         & 0.12          & 0.00         \\
                                                     &                                   & BigPatent    & 0.29    & 0.66          & 0.71         & 0.08          & -0.04        \\
\cmidrule{2-8}
                                                     & \multirow{3}{1.5cm}{sentence-delete}  & Billsum      & 0.55    & 0.75          & 0.74         & 0.26          & 0.06         \\
                                                     &                                   & ArXiv        & 0.47    & 0.69          & 0.61         & 0.34          & 0.11         \\
                                                     &                                   & BigPatent    & 0.50    & 0.79          & 0.72         & 0.35          & 0.16         \\
\midrule
\multirow{4}{*}{Baselines}                           & \multicolumn{2}{c}{Blanc-tune}                   & 0.51    & 0.73          & 0.68         & 0.33          & 0.13         \\
                                                     & \multicolumn{2}{c}{summaQA-F1}                   & 0.34    & 0.59          & 0.55         & 0.21          & 0.09         \\
                                                     & \multicolumn{2}{c}{SummaQA-CFD}                  & 0.33    & 0.60          & 0.52         & 0.25          & 0.06         \\
                                                     & \multicolumn{2}{c}{Supert}                       & 0.55    & 0.77          & 0.77         & 0.27          & 0.09         \\
\midrule
\multirow{9}{*}{Upper bounds}                        & \multicolumn{2}{c}{R-1}                          & 0.55    & 0.26          & 0.25         & 0.66          & 0.26         \\
                                                     & \multicolumn{2}{c}{R-2}                          & 0.69    & 0.03          & 0.03         & 0.59          & 0.24         \\
                                                     & \multicolumn{2}{c}{R-L}                          & 0.48    & 0.14          & 0.13         & 0.62          & 0.25         \\
                                                     & \multicolumn{2}{c}{MoverScore}                   & 0.68    & 0.06          & 0.09         & 0.51          & 0.20         \\
                                                     & \multicolumn{2}{c}{BertScore}                    & 0.65    & 0.29          & 0.28         & 0.61          & 0.24         \\
                                                     & \multicolumn{2}{c}{BLEU}                         & 0.62    & -0.14         & -0.10        & 0.32          & 0.15         \\
                                                     & \multicolumn{2}{c}{METEOR}                       & 0.71    & 0.08          & 0.09         & 0.67          & 0.28         \\
                                                     & \multicolumn{2}{c}{S3\_pyr}                      & 0.76    & 0.11          & 0.10         & 0.67          & 0.28         \\
                                                     & \multicolumn{2}{c}{S3\_resp}                     & 0.76    & 0.04          & 0.04         & 0.65          & 0.28         \\
\midrule
\multicolumn{3}{c}{Our best over baseline best (\%)}                                                    & 0.15    & 2.75          & 1.37         & 7.12          & 28.53        \\
\midrule
\multirow{6}{1.5cm}{Our average absolute deviation (\%)} & \multicolumn{2}{c}{crosspair}                    & 2.41    & 0.42          & 1.02         & 0.00          & 0.97         \\
                                                     & \multicolumn{2}{c}{sentence-delete}              & 2.85    & 3.53          & 5.27         & 3.68          & 3.68         \\
                                                     & \multicolumn{2}{c}{sentence-replace}             & 9.43    & 2.74          & 3.43         & 5.65          & 5.04         \\
                                                     & \multicolumn{2}{c}{word-insert}                     & 2.74    & 1.92          & 1.75         & 3.80          & 0.57         \\
                                                     & \multicolumn{2}{c}{word-delete}                  & 0.42    & 2.78          & 2.97         & 2.35          & 1.25         \\
                                                     & \multicolumn{2}{c}{word-replace}                 & 2.85    & 2.49          & 2.57         & 1.74          & 1.46        \\
\bottomrule
\end{tabular}
\end{table}

\begin{table}[!htb]
\setlength{\tabcolsep}{0.1em} 
\centering
\scriptsize
\caption{Full results for \textbf{Pearson}'s correlation on \textbf{linguistic} aspects.}\label{tab:ling_r}
\begin{tabular}{cccc|cc}
\toprule
                                                     &  \multirow{2}{*}{Mutation}                                 & Training             & TAC2010    & \multicolumn{2}{c}{Newsroom} \\
                                                     &                   &  set & Linguistic & COH           & FLU          \\
\midrule
\multirow{18}{1.5cm}{Our Approach}                       & \multirow{3}{1.5cm}{crosspair}        & Billsum      & 0.39       & 0.52          & 0.46         \\
                                                     &                                   & ArXiv        & 0.39       & 0.50          & 0.44         \\
                                                     &                                   & BigPatent    & 0.40       & 0.51          & 0.44         \\
\cmidrule{2-6}
                                                     & \multirow{3}{1.5cm}{sentence-delete}  & Billsum      & 0.48       & 0.61          & 0.55         \\
                                                     &                                   & ArXiv        & 0.39       & 0.56          & 0.50         \\
                                                     &                                   & BigPatent    & 0.43       & 0.65          & 0.57         \\
\cmidrule{2-6}
                                                     & \multirow{3}{1.5cm}{sentence-replace} & Billsum      & 0.43       & 0.52          & 0.44         \\
                                                     &                                   & ArXiv        & 0.21       & 0.48          & 0.42         \\
                                                     &                                   & BigPatent    & 0.39       & 0.45          & 0.38         \\
\cmidrule{2-6}
                                                     & \multirow{3}{1.5cm}{word-insert}         & Billsum      & 0.45       & 0.60          & 0.56         \\
                                                     &                                   & ArXiv        & 0.35       & 0.56          & 0.52         \\
                                                     &                                   & BigPatent    & 0.32       & 0.52          & 0.46         \\
\cmidrule{2-6}
                                                     & \multirow{3}{1.5cm}{word-replace}     & Billsum      & 0.47       & 0.61          & 0.58         \\
                                                     &                                   & ArXiv        & 0.35       & 0.56          & 0.53         \\
                                                     &                                   & BigPatent    & 0.33       & 0.53          & 0.48         \\
\cmidrule{2-6}
                                                     & \multirow{3}{1.5cm}{word-delete}      & Billsum      & 0.56       & 0.69          & 0.67         \\
                                                     &                                   & ArXiv        & 0.51       & 0.67          & 0.66         \\
                                                     &                                   & BigPatent    & 0.49       & 0.66          & 0.64         \\
\midrule
\multirow{4}{*}{Baselines}                           & \multicolumn{2}{c}{Blanc-tune}                   & 0.42       & 0.62          & 0.59         \\
                                                     & \multicolumn{2}{c}{summaQA-F1}                   & 0.29       & 0.51          & 0.47         \\
                                                     & \multicolumn{2}{c}{SummaQA-CFD}                  & 0.21       & 0.48          & 0.43         \\
                                                     & \multicolumn{2}{c}{Supert}                       & 0.46       & 0.65          & 0.58         \\
\midrule
\multirow{9}{*}{Upper bounds}                        & \multicolumn{2}{c}{R-1}                          & 0.27       & 0.17          & 0.14         \\
                                                     & \multicolumn{2}{c}{R-2}                          & 0.40       & -0.02         & -0.02        \\
                                                     & \multicolumn{2}{c}{R-L}                          & 0.18       & 0.07          & 0.06         \\
                                                     & \multicolumn{2}{c}{MoverScore}                   & 0.43       & 0.02          & 0.00         \\
                                                     & \multicolumn{2}{c}{BertScore}                    & 0.50       & 0.21          & 0.17         \\
                                                     & \multicolumn{2}{c}{BLEU}                         & 0.36       & -0.14         & -0.12        \\
                                                     & \multicolumn{2}{c}{METEOR}                       & 0.46       & 0.03          & 0.02         \\
                                                     & \multicolumn{2}{c}{S3\_pyr}                      & 0.45       & 0.04          & 0.03         \\
                                                     & \multicolumn{2}{c}{S3\_resp}                     & 0.44       & -0.01         & -0.02        \\
\midrule
\multicolumn{3}{c}{Our best over baseline best (\%)}                                                    & 21.28      & 6.71          & 13.50        \\
\midrule
\multirow{6}{1.5cm}{Our average absolute deviation (\%)} & \multicolumn{2}{c}{crosspair}                    & 0.43       & 0.64          & 0.93         \\
                                                     & \multicolumn{2}{c}{sentence-delete}              & 3.01       & 3.20          & 2.65         \\
                                                     & \multicolumn{2}{c}{sentence-replace}             & 8.89       & 2.51          & 2.39         \\
                                                     & \multicolumn{2}{c}{word-insert}                     & 5.29       & 2.86          & 3.35         \\
                                                     & \multicolumn{2}{c}{word-delete}                  & 2.56       & 1.27          & 0.98         \\
                                                     & \multicolumn{2}{c}{word-replace}                 & 6.02       & 2.88          & 3.25        \\
\bottomrule
\end{tabular}
\end{table}

\end{document}